\documentclass{article}

\usepackage[preprint]{neurips_2021}

\usepackage[utf8]{inputenc} 
\usepackage[T1]{fontenc}    
\usepackage{hyperref}       
\usepackage{url}            
\usepackage{booktabs}       
\usepackage{amsfonts}       
\usepackage{nicefrac}       
\usepackage{microtype}      
\usepackage{xcolor}         
\usepackage{float}          

\usepackage{todonotes}
\usepackage{amsmath}    
\usepackage{amssymb}    

\let\emptyset\varnothing

\newcommand*{\tran}{^{\mkern-1.5mu\mathsf{T}}}

\usepackage{multirow}   
\usepackage{subcaption} 
\usepackage{multicol}

\usepackage{bm}
\usepackage[capitalize, noabbrev]{cleveref}
\crefname{equation}{}{}
\crefname{subsection}{Subsection}{Subsections}
\Crefname{subsection}{Subsection}{Subsections}
\crefname{subsubsection}{Subsubsection}{Subsubsections}
\Crefname{subsubsection}{Subsubsection}{Subsubsections}

\usepackage{xspace}

\newcommand*{\E}[1]{\mathbb{E}\left[#1\right]}
\newcommand*{\fuzzyand}[2]{#1\wedge#2}
\newcommand*{\brackets}[1]{\begin{bmatrix}#1\end{bmatrix}}
\newcommand*{\ie}{\textit{i.e.}\xspace}

\usepackage{changepage}

\usepackage{algorithm}
\usepackage[noend]{algpseudocode}

\makeatletter
\def\BState{\State\hskip-\ALG@thistlm}
\makeatother

\algnewcommand\algorithmicinput{\textbf{yield }}
\algnewcommand\Yield{\item[\algorithmicinput]}

\title{Intrinsic Rewards from Self-Organizing Feature Maps for Exploration in Reinforcement Learning}

\author{Marius Lindegaard
\thanks{Signifies equal contribution} \ \thanks{Massachusetts Institute of Technology, Norwegian University of Science and Technology}
\ \thanks{Corresponding author: \href{mailto:lindegrd@mit.edu}{lindegrd@mit.edu}}
\And Hjalmar Jacob Vinje
\footnotemark[1] \ \footnotemark[2]
\And Odin Aleksander Severinsen
\footnotemark[1] \ \footnotemark[2]
}

\date{Dated: 2022}

\DeclareMathOperator*{\argmax}{arg\,max}

\begin{document}
\maketitle
\begin{abstract}
We introduce an exploration bonus for deep reinforcement learning methods calculated using self-organising feature maps. Our method uses adaptive resonance theory providing online, unsupervised clustering to quantify the novelty of a state. This heuristic is used to add an intrinsic reward to the extrinsic reward signal for then to optimize the agent to maximize the sum of these two rewards.
We find that this method was able to play the game Ordeal
at a human level
after a comparable number of training epochs to ICM \citep{self_supervised_ICM}. Agents augmented with RND \citep{RND_2018} were unable to achieve the same level of performance in our space of hyperparameters.
\end{abstract}
\section{Introduction}

A central challenge in reinforcement learning (RL) is the tradeoff between exploration and exploitation.
Often times, the reward function, which will ultimately be exploited, does not provide any motivation for exploration.
Lacking motivation to explore, RL agents are often prone to getting stuck in local minima.

The exploration problem is particularly prevalent in environments with sparse rewards, complex state representations, or requiring complicated high-level decisions consisting of several actions concatenated through time.
The use of \emph{intrinsic motivation}, shown to remedy some of these problems \citep{survey_2019}, assigns rewards to actions which by some heuristic explore the environment. With exploration bonuses the policy is still optimized to maximize rewards.
This allows implementation of exploration bonuses by simple modifications to nothing but the reward signal. In this paper, we explore a method of calculating intrinsic reward using a form of online \textit{self-organizing feature maps} (SOFMs).

Our implementation is available at 
\href{https://github.com/mariuslindegaard/curiosity_baselines}{https://github.com/mariuslindegaard/curiosity\_baselines}.

\section{Related work}

Previous methods for providing intrinsic reward include use of next-state prediction (\cite{self_supervised_ICM}, \cite{self_supervised_disagreement}) and state recollection \citep{RND_2018}.
The next-state prediction methods derive some exploration bonus $i_t$ from predicting the next state $s_{t+1}$ from the current state and action $(s_t, a_t)$, often through a feature embedding to a lower dimensional space.
The single state recollection methods use the state $s_{t+1}$ alone to assign an exploration bonus $i_t$, intuitively giving some bonus to states dissimilar to what has previously been seen.


Our work centeres around online algorithms of \emph{self-organizing feature maps} for learning structure from data.
Several SOFMs satisfy all the properties required for calculating exploration bonuses.
We consider methods of \emph{adaptive resonance theory} (ART) \citep{grossberg_1976} inspired by computational neuroscience, in particular \emph{Fuzzy ART} \citep{CARPENTER1991759}.

We are using the code base introduced in \citep{curiosity_baseline}, which is a framework for implementing curiosity driven agents in game environments.

\section{Motivation}\label{sec:motivation}
Intuitively, the exploration bonuses using next-state prediction and state recollection are in many properties similar to online unsupervised anomaly detection. This is a domain in which ART has shown promising results, being the current state of the art on the Numenta Anomaly Benchmark \citep{ART_2017_anomalydet}.
We suspect this performance translates to giving relevant exploration bonuses for novel states in reinforcement learning.
Therefore, we propose replacing the mechanism for providing exploration bonuses by with SOFMs. Since ART has performed best in class for anomaly detection, we hypothesize it can be used for more sample efficient exploration in game environments. 

As mentioned, SOFMs has all the important functional properties of the exploration bonus in the RL algorithms. These are (a) online learning of data structure, (b) being self-/unsupervised and (c) including a measure of how different the given input is from previously seen data.
This final measure (c) is used as a reward for sampling data differing significantly from previous samples.
This is the heuristic rewarding exploration into previously unseen states.

Intrinsic motivation has been crucial in solving complex RL tasks in exploration heavy environments \citep{RND_2018}, and improving the methods for exploration bonus will allow solving an even wider array of tasks in reinforcement learning. Alternative approaches to the previously explored methods could provide new insight into the reasons for the success and failures of exploration bonus based methods.

%
%
\section{Theoretical background}

The following section will give a brief introduction to the overall problem of learning that we seek to solve in reinforcement learning. Then, the \emph{Proximal Policy Optimization (PPO)} algorithm that is used for policy optimization. Lastly, the self-organizing feature map methods ART are introduced, together with how intrinsic rewards are computed.

\subsection{Reinforcement learning algorithms}


Fundamentally, \emph{reinforcement learning} is about mapping a trajectory or observations to actions in order to maximize a reward signal given by the environment. 


Following this, we wish to find a policy that maximizes the discounted expected return, denoted by 
\begin{equation}
    R^{\gamma}(\tau) = \E{\sum_{i=1}^T\gamma^{i-1}r_i(s_i, a_i)}
\end{equation}
where $r_i$ denotes the reward from timestep $i$ by doing action $a_i$ in state $s_i$, $\tau$ denotes the trajectory of the agent and is given as $\tau=\{s_0, a_0, r_0, \dots, s_T, a_T, r_T\}$ and $\gamma\in [0, 1]$ is a given discount factor.

This paper considers \emph{policy gradient methods} where we learn a stochastic policy $\pi_{\theta}(a_t|s_t)$ that defines a probability distribution over actions $a_t$ given the current state $s_t$ parameterized by $\theta\in\mathbb{R}^m$. Actions are then chosen by sampling from this distribution.

Theoretically, we can find the optimal policy based on some training set of samples from the environment and then solve the optimization problem
\begin{equation}\label{eq:theory:theta_argmin}
    \theta^* = \argmax_{\theta} J(\theta)
\end{equation}
for a suitable maximization target given by $J(\theta)$.

\subsection{Proximal Policy Optimization}

There are multiple considerations to take to robustly converge to a solution, most notably variance reduction of the estimated gradient that has to be computed. In order to maximize policy improvement we use the sequential optimization objective
\begin{align}
    \theta_{t+1} &= \theta_t + \Delta \theta_t, \label{eq:policy_improvement_optimize} \\
    \Delta \theta_t &= \argmax_d J(\theta_t+d) - J(\theta_t),
\end{align}
where $d$ denotes a change to the parameters from one iteration to the next. It can be shown \citep{kakade_langford_2002} that the objective can be rewritten as
\begin{align}
    J(\theta + d) - J(\theta) &= \mathbb{E}_{\tau\sim\pi_{\theta + d}}\left[\sum_{t=1}^T\gamma^{t-1}A^{\pi_{\theta}}(s_t, a_t)\right] \\
    &= \mathbb{E}_{\substack{s\sim\pi_{\theta + d}, \\ a\sim\pi_{\theta}}}\left[\frac{\pi_{\theta+d}(a|s)}{\pi_{\theta}(a|s)}A^{\pi_{\theta}}(s,a)\right].\label{eq:objective_J}
\end{align}
where the \emph{advantage}
\begin{equation}
    A^{\pi_{\theta}}(s_t, a_t) = R^{\gamma}(\tau_{t:T}) - b(s_t)
\end{equation}
and where $\tau_{t:T}$ denotes the subtrajectory of the total trajectory only considering the timesteps from $t$ to $T$ and $b(s_t)$ is some bias term dependent on the state $s_t$.

In addition to optimizing the objective in \eqref{eq:objective_J}, it is common to constrain the KL-divergence between the two policies $\pi_{\theta + d}$ and $\pi_{\theta}$. 
Simplifying the KL-convergence constraint, \emph{Proximal Policy Optimization} (PPO) replaces the KL-divergence constraint with a clipping of the ratio $\frac{\pi_{\theta+d}(a|s)}{\pi_{\theta}(a|s)}$ between $1-\epsilon$ and $1 + \epsilon$ for some $\epsilon$ to penalize big changes between the two policies. Thus, the final objective takes the form
\begin{equation}
    J(\theta + d) - J(\theta) = \mathbb{E}_{\substack{s\sim\pi_{\theta + d}, \\ a\sim\pi_{\theta}}}\left[\min\left\{\frac{\pi_{\theta+d}(a|s)}{\pi_{\theta}(a|s)}A^{\pi_{\theta}}(s,a), \mathrm{clip}\left(\frac{\pi_{\theta+d}(a|s)}{\pi_{\theta}(a|s)}, 1-\epsilon, 1+
    \epsilon\right)A^{\pi_{\theta}}(s,a)\right\}\right].
\end{equation}

We will in our experiments be using PPO as the underlying algorithm and let the reward $r_t = e_t \to r_t = e_t + i_t$ where $e_t$ are the extrinsic rewards provided by the environment and $i_t$ are exploration bonuses generated by observing the current state and action. Our experiments will revolve around implementing and evaluating two new methods for providing exploration bonuses $i_t$.

\subsection{Exploration bonuses} \label{sec:Exploration bonuses}

The use of exploration bonuses in RL methods is intuitively meant to encourage exploration even when the extrinsic reward $e_t$ is sparse in the environment.
The agent $\pi$ is trained to maximize the total reward $r_t = e_t + i_t$ where the \emph{intrinsic reward} $i_t$ is some exploration bonus.
Using this exploration bonus the agents can learn tasks in environments with sparser rewards than methods based purely on extrinsic rewards.

There are several methods for assigning exploration bonuses $i_t$. Recent methods demonstrating success include ICM and RND, which here are also used as baselines. The ICM method computes the exploration bonus by first encoding the states $s_t$ and $s_{t+1}$ as the feaatures $\phi(s_t)$ and $\phi(s_{t+1})$, for then to compute a prediction $\hat{\phi}(s_{t+1})$ of the next state in feature space with $\phi(s_t)$ and action $a_t$. The exploration bonus is then given as the prediction error between the next states in feature space \citep{self_supervised_ICM}. RND uses a prediction network $f_{\theta}(x)$ in conjunction with a fixed, random target network $\hat{f}_{\theta}(x)$ to compute exploration bonuses. When a new observation $x$ is received, it is propagated through both networks, and the error between the two outputs is used as exploration bonus. The prediction network is then trained to give a similar output as the target network \citep{RND_2018}. For a more thorough discussion on exploration bonus based methods, the reader is referred to \citep{Understanding_bonus}.

\subsection{Adaptive Resonance Theory}




The field of adaptive resonance theory was first introduced by Stephen Grossberg and Gail Carpenter \citep{grossberg_1976,carpenter_grossberg_1987}, and is proposed as a more biologically consistent theory for how the human brain does learning in a constantly evolving environment \citep{art_survey_2019}. As ART is a SOFM, it benefits from the properties listed in \cref{sec:motivation}. However, the perhaps most useful property of ART compared to other, more traditional neural network models is that it addresses the \emph{stability-plasticity dilemma} \citep{carpenter_grossberg_1987a,grossberg_1980} which is about the balance between designing a system that is able to remember previous experiences (stability) while also being able to make new memories based on novel experiences (plasticity). While ART is able to remember previous experiences and also learn new ones, modern, neural networks are in general not, a phenomena called \emph{catastrophic forgetting} \citep{MCCLOSKEY1989109}.

The following treatment of Fuzzy ART is based on \cite{art_survey_2019}.



\subsubsection{Fuzzy ART}

The used implementation for testing is the \emph{Fuzzy ART} \cite{CARPENTER1991759}, which generalizes the original ART formulation to accept real-valued features by using operators from fuzzy set theory.

First, an augmentation of the input $\bm{\mathsf{x}}\in[0, 1]^d$ is done to prevent proliferation of categories due to to weight erosion \cite{CARPENTER19971473}. Thus, the new, augmented input $\bm{x}$ takes the form
\begin{equation}
    \bm{x} = \begin{bmatrix}\bm{\mathsf{x}} \\ \bm{1} - \bm{\mathsf{x}}\end{bmatrix},\quad \bm{x} \in [0, 1]^{2d}.
\end{equation}
The long-term memory (LTM) in Fuzzy ART is given by a discrete, finite set of weight vectors 
\begin{equation}\label{eq:fuzzy_and:LTM}
\mathcal{W}=\bigcup_{j\in\mathcal{C}}\{\bm{w}_j\}, 
\end{equation}
where $\mathcal{C}= \{1,\dots,C\}$ is the index set over all $C$ categories stored in the LTM so far and $\bm{w}\in[0, 1]^{2d}$. When a new input is presented to Fuzzy ART, the \emph{activation} $T_j$ of this input is computed for each node $\bm{w}_j$ in the LTM as
\begin{equation}
    T_j = \frac{||\bm{x}\wedge \bm{w}_j||_1}{\alpha + ||\bm{w}_j||_1},\quad\forall j\in\mathcal{C}\label{eq:fuzzy_art:activation},
\end{equation}
where $\fuzzyand{\bm{u}}{\bm{v}}$ denotes the \emph{fuzzy AND} of the two vectors $\bm{u},\bm{v}\in\mathbb{R}^n$ and is defined as the vector of element-wise $\min(\cdot)$ of $\bm{u}$ and $\bm{v}$ such that
\begin{equation}
    \fuzzyand{\bm{u}}{\bm{v}} = \brackets{\min(u_1, v_1) \\ \vdots \\ \min(u_n, v_n)},
\end{equation}
the 1-norm $||\bm{u}||_1$ for $\bm{u}\in\mathbb{R}^n$ is
\begin{equation}
    ||\bm{u}||_1 = \sum_{i=1}^n |u_i|,
\end{equation}
and $\alpha>0$ is a regularizing hyperparameter that penalizes large weights. For the next step, define the ordered, sorted index set
$\mathcal{C}^* = \{J~|~J\in\mathcal{C},~T_{J-1} \geq T_J \geq T_{J+1}\}$,
\ie, with indices in order of highest activation to lowest. The \emph{vigilance test} is conducted to check for resonance in each node from the input by computing the \emph{vigilance}
\begin{equation}\label{eq:fuzzy_and:vigilance}
    M_J = \frac{||\bm{x}\wedge \bm{w}_J||_1}{||\bm{x}||_1},\quad \forall J\in\mathcal{C}^*  
\end{equation}
and then comparing it to the \emph{vigilance threshold} $\rho\in[0, 1]$
\begin{equation}\label{eq:fuzzy_art:vigiliance_thresh}
    M_J \geq \rho
\end{equation}
where the winning node is the first node $J$ to satisfy \cref{eq:fuzzy_art:vigiliance_thresh}. In case no node satisfy \cref{eq:fuzzy_art:vigiliance_thresh}, a new category is initialized with all weights equal to $1$, such that
\begin{align}
    \bm{w}_{C+1} &= \bm{1}, \\
    \mathcal{W} &\gets \mathcal{W}\cup\{\bm{w}_{C+1}\}, \\
    J &= C + 1.
\end{align}

The last step to Fuzzy ART is the learning step, where the weight $\bm{w}_{J}$ of the winning node $J$ is updated according to
\begin{equation}\label{eq:fuzzy_art:learning}
    \bm{w}_J \gets (1 - \beta)\bm{w}_J + \beta(\fuzzyand{\bm{x}}{\bm{w}_J})
\end{equation}
where $\beta$ is the learning rate. The full algorithm for joint learning and classification of one feature $\bm{x}$ can be found in \cref{alg:FuzzyART}. 

\begin{algorithm}
\caption{Fuzzy ART learning and classification.}\label{alg:FuzzyART}
\begin{algorithmic}[1]
\Procedure{FuzzyART}{$\bm{x}$, $\mathcal{W}$}
    \State $\bm{T}$ $\gets \left[T_j\text{ for }j\text{ in }\mathcal{C}\right]$ \Comment{Compute activation of each node for input with \cref{eq:fuzzy_art:activation}}
    \State $J_{\mathrm{winner}} \gets C + 1$
    \State $\mathcal{C}^* \gets \textsc{SortedSet}(\mathcal{C}, \bm{T})$
    \For{$J\in\mathcal{C}^*$}\Comment{Loop over ordered indices}
        \If{$M_J\geq\rho$} \Comment{Vigilance test based on \cref{eq:fuzzy_and:vigilance} and \cref{eq:fuzzy_art:vigiliance_thresh}}
            \State $J_{\mathrm{winner}}\gets J$
            \State \textbf{break}
        \EndIf
    \EndFor
    \If{$J_{\mathrm{winner}} = C + 1$} \Comment{No node resonated with input, initialize new category}
    \State $\bm{w}_{C+1} \gets \bm{1}$
    \State $\mathcal{W} \gets \mathcal{W}\cup\{\bm{w}_{C+1}\}$
    \EndIf
    \State $\bm{w}_{J_\mathrm{winner}} \gets (1 - \beta)\bm{w}_{J_\mathrm{winner}} + \beta(\fuzzyand{\bm{x}}{\bm{w}_{J_\mathrm{winner}}})$ \Comment{Learning}
    \State \Return $J_{\mathrm{winner}}$
\EndProcedure
\end{algorithmic}
\end{algorithm}


\subsubsection{Batch learning and classification in Fuzzy ART}
In practice, training in Fuzzy ART is done on batches of features $\mathsf{X}=\{\bm{\mathsf{x}}_1, \dots, \bm{\mathsf{x}}_B\}$ simultaneously, which requires a slight modification of the algorithm described in \cref{alg:FuzzyART}. In this case, we do a shuffling of the features in the batch and do learning on all of them sequentially, updating the LTM $\mathcal{W}$ for each, and continuing until the weights $\bm{w}$ of the LTM converges or the number of epochs of training $n_{\mathrm{epoch}}$ surpasses a predetermined threshold $N_{\max}$. See \cref{alg:OnlineBatchFuzzyART} for steps on how to do learning of batches of features.

\begin{algorithm}[H]
\caption{Fuzzy ART for online batch learning. The full \textsc{FuzzyART} learning algorithm can be found in \cref{alg:FuzzyART}.}\label{alg:OnlineBatchFuzzyART}
\begin{algorithmic}[1]
\Procedure{OnlineBatchFuzzyART}{$\mathsf{X}$, $\mathcal{W}$}
    \State $n_{\mathrm{epoch}}\gets 0$
    \While{\textsc{NotConverged}($\mathcal{W}$) \textbf{ and } $n_{\mathrm{epoch}}<N_{\max}$}
        \State $\mathcal{J} \gets \emptyset$ \Comment{Initilize set of classifications by Fuzzy ART on features}
        \State $\mathcal{X}\gets$\textsc{Shuffle}($\mathsf{X}$)
        \For{$\bm{\mathsf{x}}\in\mathcal{X}$}
            \State $\bm{x}$ $\gets \brackets{\bm{\mathsf{x}}\tran & \bm{1}-\bm{\mathsf{x}}\tran}\tran$
            \State $J\gets$\textsc{FuzzyART}($\bm{x}, \mathcal{W}$)
            \State $\mathcal{J}\gets \mathcal{J}\cup \{J\}$ 
        \EndFor
        \State $n_{\mathrm{epoch}}\gets n_{\mathrm{epoch}}+1$
    \EndWhile
    \State \Return $\mathcal{J}$
\EndProcedure
\end{algorithmic}
\end{algorithm}

\subsubsection{Fuzzy ART input encoding}

There are two particular practical considerations to make when encoding features to be used by Fuzzy ART.
Firstly, Fuzzy ART only accepts inputs $\bm{x}\in[0, 1]^d$, and a transformation to perform such a mapping is needed in case this is not the case. Additionally, the observation space might be large, and hence it can be beneficial to represent the features with a more compact encoding for performance reasons. In the experiments, a static feature encoding \emph{head} was used to preprocess the observations for some of the ART models. We also present a \emph{headless} version (ART-HL), working directly on the $(8\times5\times5)$ binary input space of Ordeal.



\subsection{Exploration bonus specification}

As discussed, ART will be used as a curiosity model, and so a suitable function for computing the intrinsic reward $i_t$ from an observation $\bm{x}_t$ is necessary. In \cite{Bellemare_2016}, it is argued that a suitable metric for quantifying novelty in the observation is the \emph{information gain} \cite{cover_thomas_1991}, defined as the KL divergence of the \emph{prior state distribution} $\rho_t(x)$, \ie the distribution over $\bm{x}$ prior to an observation $\bm{x}_t$, from its corresponding \emph{posterior distribution} $\rho'(\bm{x})$.
In general, this quantity is often not tractable to compute. They claim, however, that the \emph{prediction gain} $PG_n$, given by $PG_t = \log\rho'_n(\bm{x}_t) - \log\rho_n(\bm{x}_t)$, is a good approximation. They then show that the inverse square-root of the \emph{pseudo-count} $\hat{N}_t(\bm{x})$ over times that state $\bm{x}$ has been visited can be used as an upper bound for the prediction gain as $PG_t \leq \hat{N}_t(\bm{x})^{-1/2}$ in which Bellemare et al. claim that an exploration bonus proportional to $\hat{N}_t(\bm{x})^{-1/2}$ will give behavior that is at least as exploratory as as an exploration bonus based on information gain. 

By denoting Fuzzy ART by the function $j = f(\bm{x})$, where $j$ is the classification of $\bm{x}$ as computed by Fuzzy ART, and introducing the \emph{category count function} $N(j)$, which returns the number of times a category $j$ has been seen by ART, a natural choice for computing the intrinsic reward $i$ is then
\begin{equation}
    i_t(\bm{x}_t) = \frac{k}{\sqrt{N(f(\bm{x}_t))}} \quad k>0
\end{equation}
for an observation $\bm{x}_t$, which is used in the experiments that follow.
 






\section{Experiments and results}

The following section will present the experimental setup and environment used to train and test the agents, together with a discussion on the results.

\subsection{Experimental setup}
\begin{figure}[H]
    \centering
    \includegraphics[height = 50mm]{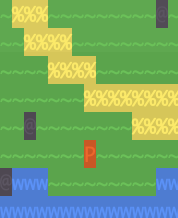}
    \includegraphics[height = 50mm]{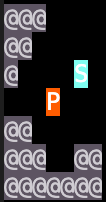}
    \caption{Visualization of the game environments: Ordeal\newline Left: Starting point. Right: Cave with sword}
    \label{fig:ordeal}
\end{figure}

To test the performance of our proposed methods we have been training our agents using Deepmind's \textit{Pycolab game engine}
\footnote{\href{https://github.com/deepmind/pycolab}{https://github.com/deepmind/pycolab}}
for evaluating the performance of our methods. The specific game we have used is called Ordeal and can be seen in \cref{fig:ordeal}. The goal of the game is to move to a cave to pick up the sword, which gives you one point. Then you should go back to another cave with a duck that approaches you and when you make contact, you kill it and get another point. However, if you did not pick up the sword first, the duck kills you and you get -1 point. The possible outcomes of the games are therefore as seen in \cref{tab:ordeal}
\begin{multicols}{2}
\hspace{0pt}
\vfill
\begin{table}[H]
    \centering
    \begin{tabular}{c|c}
    \toprule
        Game play & Points \\ \hline
        Go straight to room with duck & -1 \\
        Do nothing & 0 \\
        Find sword but not duck & 1 \\
        Find sword then duck & 2 \\
         \bottomrule
    \end{tabular}
    \caption{Outcomes in Ordeal}
    \label{tab:ordeal}
\end{table}
\vfill
\hspace{0pt}
\end{multicols}

The fact that the agent needs to go into a specific room to get something, before entering a second room is similar to the Atari game Montezuma's revenge. That game was solved with intrinsic reward from RND \citep{RND_2018} but not by traditional DQN \citep{Mnih2015HumanlevelCT}. The similarity is why we wanted to use Ordeal as test our intrinsic reward method, as we would test for similar concepts, but in a simpler environment.

As a success criterion we will use the total extrinsic rewards on this game, compared to other methods for giving exploration bonuses for RL agents.

The state space is the information seen on the screen around the agent. Specifically this is a $(8\times5\times5)$ one hot encoding for each of the 8 types of squares that can appear on screen. The action space is simply moving up, down, left and right, in addition to being able to quit the game.

\subsection{Quantitative results}

\begin{figure}
    \centering
    \includegraphics[height = 60mm]{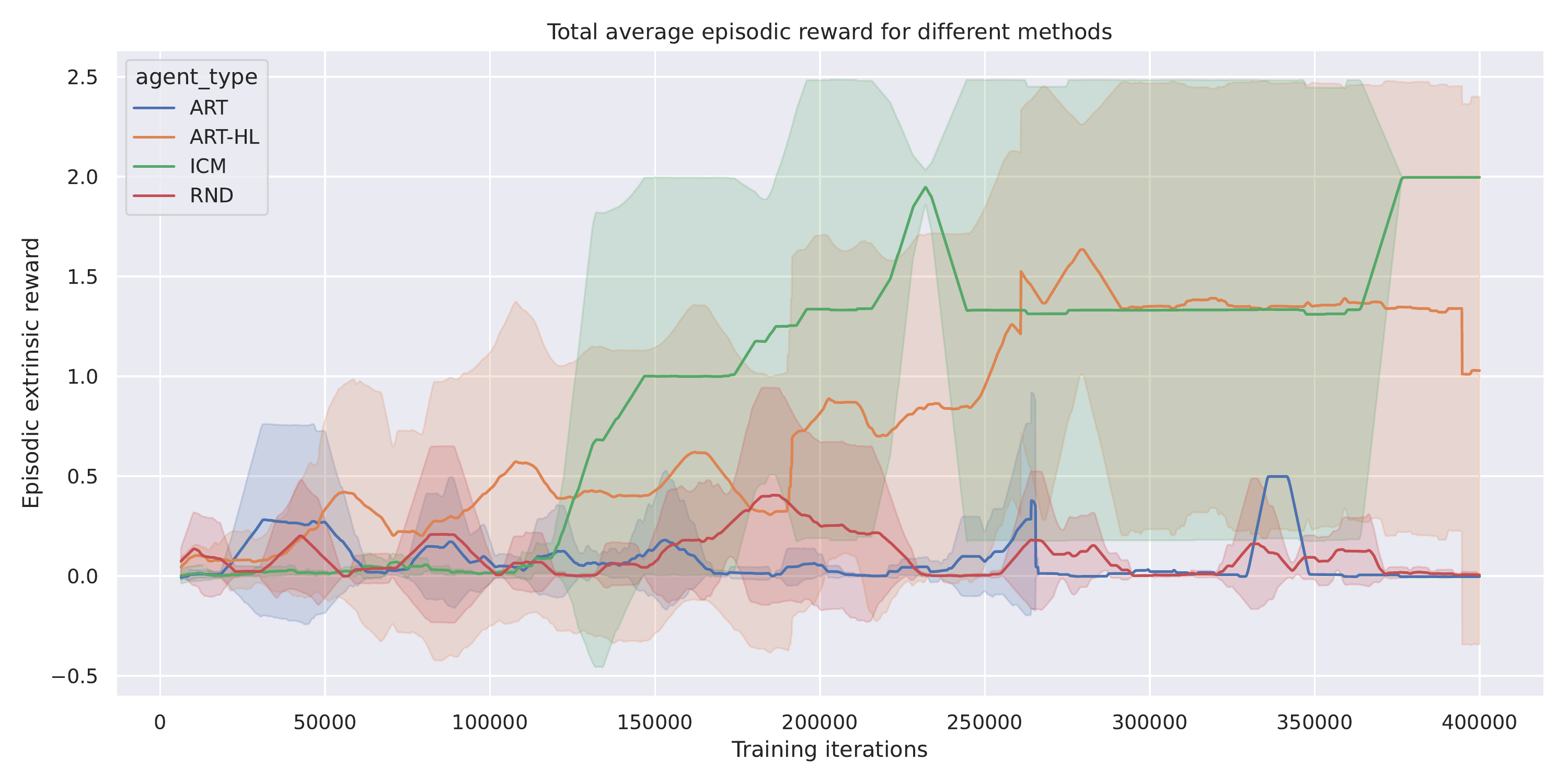}
    \caption{Performance over training iterations for different intrinsic reward algorithms. The ART-HL method is the headless ART, ART references ART with the static feature encoding head, RND and ICM referencing the methods mentioned in \cref{sec:Exploration bonuses}. The shaded area shows the standard deviation in the results}
    \label{fig:performance1}
\end{figure}

Using the different methods for generating intrinsic rewards, we simulate several runs on Ordeal evaluating the average episodic intrinsic rewards. Our results are presented in \cref{fig:performance1}.
The figure shows that after 400k iterations, that ICM is able to converge to the solution in all cases. This makes it the best performing method. The headless ART is able to converge to the optima in some cases, on average achieving a score of around 1. None of the runs with ART with a dimension reducing head or RND for intrinsic reward are able to achieve significant success on the Ordeal environment.

As seen by the significant standard deviation on ART-HL and ICM, there is stochasticity in whether the algorithms reach the optimum score after a given set of training episodes.
Continued training appears to increase the probability that convergence occurs for both successful methods, as evidenced by the increasing average score.

\subsection{Qualitative analysis}\label{sec:qualitative_analysis}
From visualizing our agent behaviour through game play, we can see that it goes in a straight line to the room with the sword and then takes what looks like the shortest path to the room with the duck. The agent trained using ART therefore exhibits desired performance at an optimal or close to optimal level. A video of the ART agent solving the game can be found here. \footnote{
\href{https://drive.google.com/drive/folders/1NT1lFBW56kIZcAWlMqHXah3dwfOmHeYn?usp=sharing}{https://drive.google.com/drive/folders/1NT1lFBW56kIZcAWlMqHXah3dwfOmHeYn?usp=sharing}
}

\begin{figure}
    \centering
    \includegraphics[height =80mm]{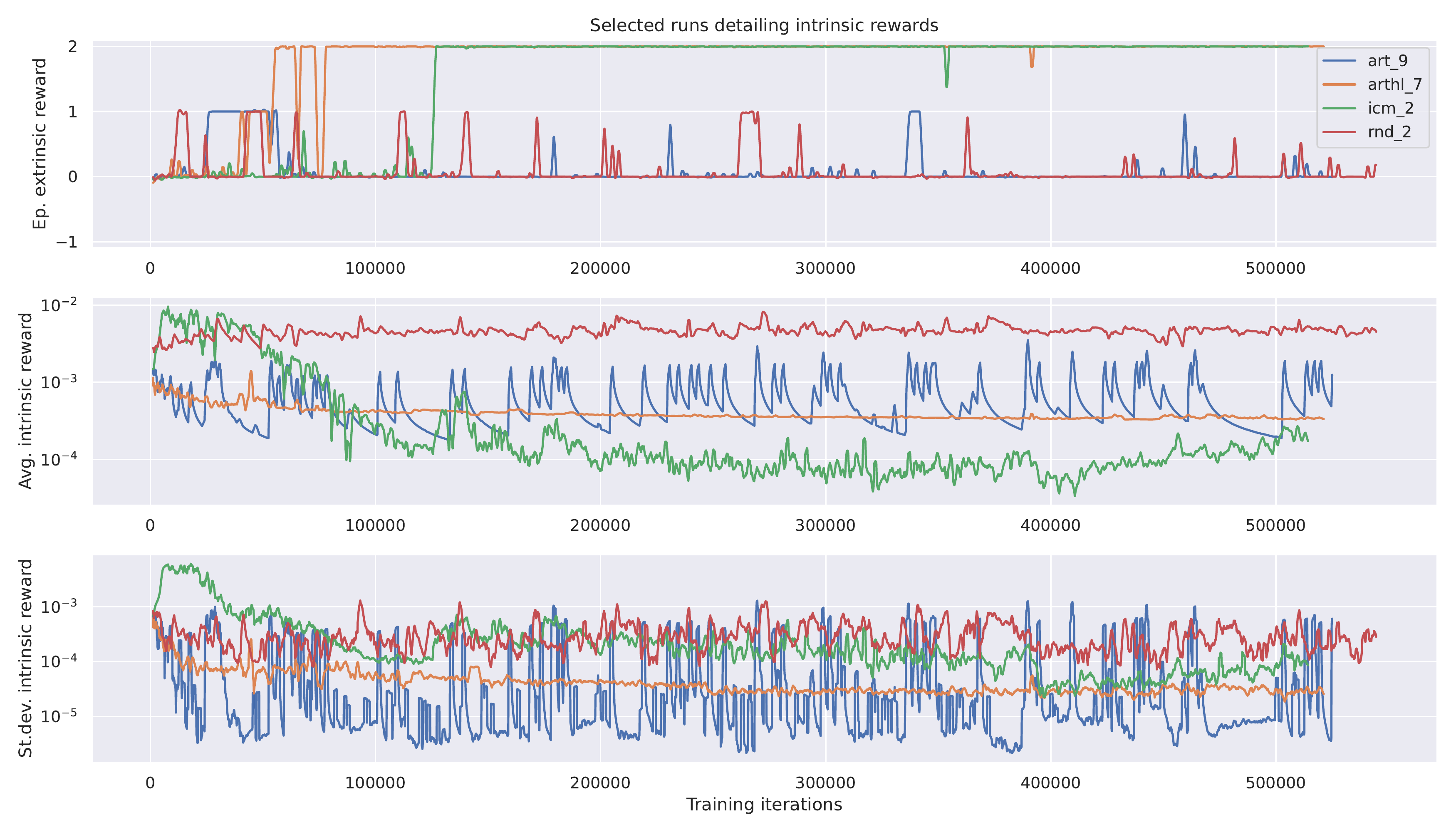}
    \caption{Performance over training iterations for different individual agents. The labelling is as in figure \ref{fig:performance1}, with lowercase and index indicating that these are individual agent runs.}
    \label{fig:performance}
\end{figure}

In \cref{fig:performance}, we can see that while RND is the first agent to get one point, it never solves the whole game to achieve 2 points.
The first agent to do this is ART. The intrinsic reward can be seen as the orange line in the second graph. We can see some desired behavior, like the fact that when the agent consistently finds the sword, and therefore moves up to one point in extrinsic reward, it coincides with a spike in the intrinsic reward. This shows a correlation between finding a new room and intrinsic reward. 
However, the jump from one to two, where it learns to go to the duck afterwards, is not associated with the same spike, so is not "rewarded" intrinsically. This might be because the second room is to similar to the first one.

For these typical runs we note the difference in average intrinsic rewards for the different methods.
While all agens have comparable variance in intrinsic reward, RND seems to have a particularly high average intrinsic reward. We hypothesise that this contributes to the fact that this method never converges toward the two-point extrinsic reward optimum, as the PPO agent rather optimizes based on gaining more intrinsic reward and choosing not to end the game.
In numbers, the problem lies in that for the maximal $500$ steps per episode  (See figure \ref{fig:rnd_performance} in Appendix \ref{app:rnd_performance}) and average intrinsic reward hovering around $0.005$ the total intrinsic reward is around $2.5$, greater than the maximal extrinsic reward of $2$. The goals of ending the episode early by finding the sword to kill the duck versus continuing to gain intrinsic reward through a long episode results in, in the case of RND, continuing "exploration" for as long as possible.

Note also the extreme instability in average intrinsic reward for the \verb|art_9| run. This is due to new classes sporadically being created with all samples fitting in this new class having an increased intrinsic reward compared to the other states. Reducing this noise by eventually capping the number of classes is likely one of the main reasons the head-less ART method performs better than the standard ART, as capping the number of classes in this relatively small environment avoids the problem of the intrinsic reward becoming the optimizing target for the PPO agent.

Finally we mention that the agents with no intrinsic reward generally exhibited very little interesting behavior and were therefore left out for brevity and clarity. Nearly all standard agents were unable to achieve any reward signal from the environment, resulting in little to no learning.
This problem is exacerbated by the "quit" option being part of the agent action space, resulting in the agent often quitting early and not exploring the environment.

\section{Conclusions}

Using the sparse reward environment Ordeal we have evaluated several methods for assigning intrinsic rewards to promote exploration of the state space. 
Our results demonstrate the ability of ART to be used as a foundation for calculation exploration bonuses by online state clustering and counting.

While not all algorithms were able to demonstrate convergence to an optimal policy for extrinsic reward, we propose that this might be due to a problem of tuning the intrinsic vs. the extrinsic reward. We leave to future work to explore tuning this parameter further for all methods (ART, ICM \& RND based), as well as the possibility of introducing artificial decay in the intrinsic reward.

Furthermore, the robustness of this method should be evaluated by comparing it to the other approaches in more complex environments with different state spaces and reward structures.



\section*{Acknowledgments}
We wish to thank to Prof. Pulkit Agrawal and of the 6.484 teaching staff for teaching an inspiring and practical course as well as their guidance and support through the project process.


\appendix

\section{Appendix}\label{app:rnd_performance}


\subsection{}
\begin{figure}[H]
    \centering
    \includegraphics[width=120mm]{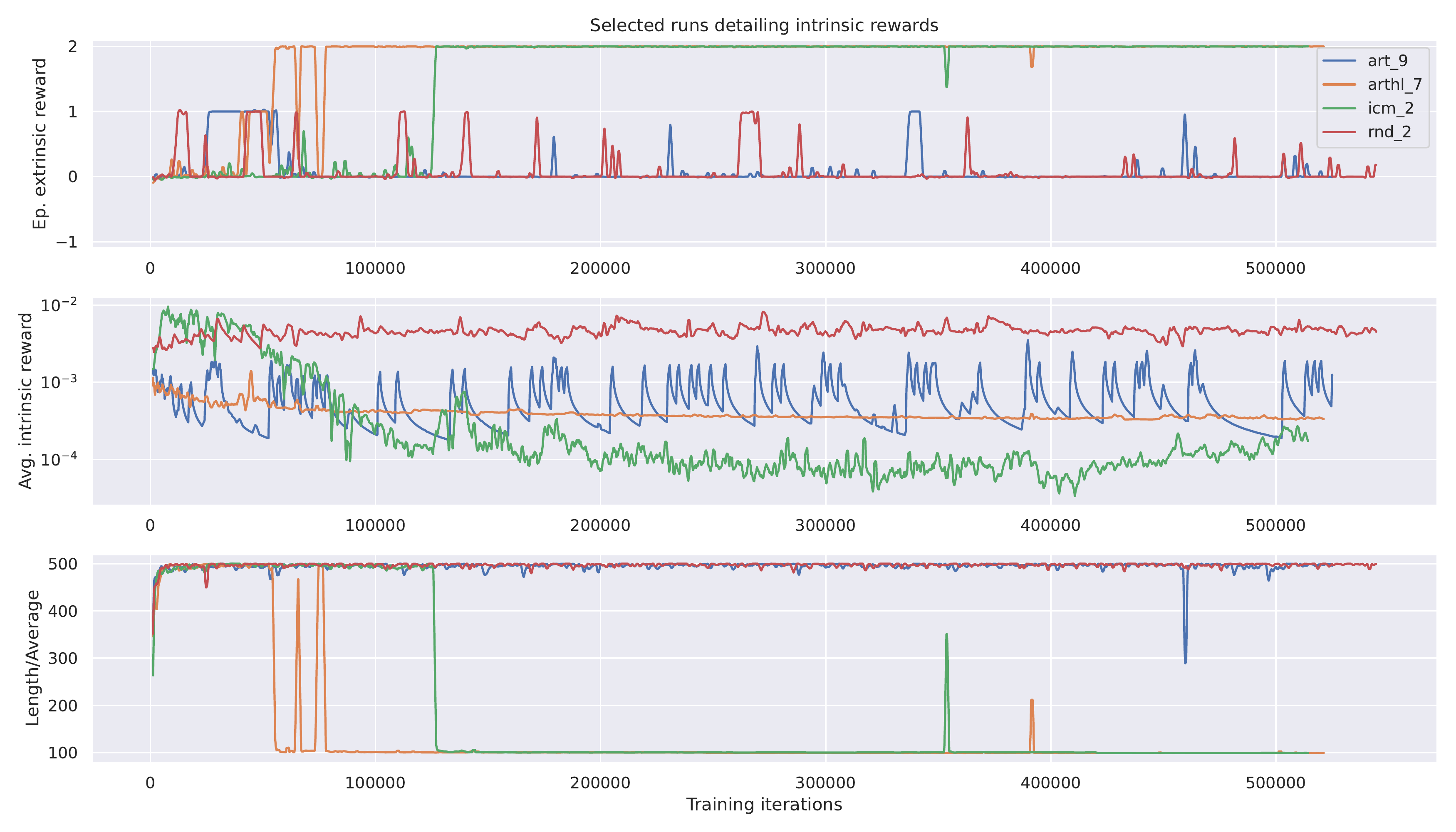}
    \caption{Selected runs with details for intrinsic rewards. Note that the two methods with the highest intrinsic rewards, ART and RND, both do not reduce their total amount of steps in the environment.}
    \label{fig:rnd_performance}
\end{figure}

\subsection{Notes on parameter tuning and compute}

For exploration using intrinsic rewards there is an obvious volatility in the results from the tuning hyperparameters. In this paper we have chosen to do some initial tuning to the algorithms, observing that the standard deviation of the intrinsic reward is approximately within an order of magnitude for the different algorithms. This can be important as the advantage function, on which PPO is trained, relies on the difference between the expected value of taking different actions.

On the other hand, as noted in section \ref{sec:qualitative_analysis}, this environment has a variable episode length which might lead to the PPO algorithm valuing to continue receiving intrinsic reward rather than optimizing the extrinsic reward.

Although we performed some tuning before running our final simulations, specifically scaling down the intrinsic reward for all algorithms, we recognize that more time tuning all relevant hyperparameters and access to more compute would likely yield better results for all algorithms.

\bibliography{references}
\end{document}